\documentclass{article}

    \PassOptionsToPackage{numbers, compress}{natbib}
 \usepackage[preprint]{neurips_2026}

\usepackage[utf8]{inputenc} 
\usepackage[T1]{fontenc}    
\usepackage{hyperref}       
\usepackage{url}            
\usepackage{booktabs}       
\usepackage{amsfonts}       
\usepackage{nicefrac}       
\usepackage{microtype}      
\usepackage{xcolor}         

\usepackage{graphicx}
\usepackage{subcaption}
\usepackage{tikz}
\usepackage{amsmath}
\usepackage{dsfont}
\usepackage{longtable}
\usepackage{tabularx,colortbl}
\usepackage{multirow}
\usepackage{soul}
\usepackage{algorithm}
\usepackage{algorithmic}
\usepackage{wrapfig}

\title{AROpt: An Optimization Method for Autoregressive Time Series Forecasting}

\author{%
  \small Zheng Li \\
  \scriptsize Department of Computer Science\\
  \scriptsize New York Institute of Technology\\
  \small New York, NY 10023 \\
  \small \texttt{zli66@nyit.edu} \\
  \And
  \small Jerry Cheng \\
  \scriptsize Department of Computer Science\\
  \scriptsize New York Institute of Technology\\
  \small New York, NY 10023 \\
  \small \texttt{jcheng18@nyit.edu} \\
  \And
  \small Huanying Helen Gu \\
  \scriptsize Department of Computer Science\\
  \scriptsize New York Institute of Technology\\
  \small New York, NY 10023 \\
  \small \texttt{hgu03@nyit.edu} \\
}

\begin{document}

\maketitle

\begin{abstract}
  Current time-series forecasting models are primarily based on transformer-style neural networks. These models achieve long-term forecasting mainly by scaling up the model size rather than through genuinely autoregressive (AR) rollout. From the perspective of large language model training, traditional time-series forecasting model training ignores the AR rollout errors accumulate heuristic. In this paper, we propose a novel training method for time-series forecasting that introduces a rollout-aware objective while preserving the original optimization target: (1) AR prediction errors should increase with the forecasting horizon. Violations of this trend are interpreted as rollout inconsistency and are softly penalized during training, and (2) the method enables models to be able to concatenate short-term AR predictions to form flexible long-term forecasts. Empirical results demonstrate that our method demonstrates consistent improvements over its standard training counterparts across multiple benchmarks. Furthermore, models trained only for short-horizon forecasting generalize to forecasting horizons over 7.5 times longer through AR rollout. Code is available at  \url{https://github.com/LizhengMathAi/AROpt}
\end{abstract}

\section{Introduction}
\begin{wrapfigure}[21]{r}{0.5\textwidth}
  \vspace{-31pt}
  \begin{center}
    \centerline{
        \begin{tikzpicture}
        \node[inner sep=0pt, outer sep=0pt] (img) {%
          \includegraphics[
            width=0.5\columnwidth,
            trim=2.5cm 16.75cm 2.5cm 1.8cm,
            clip
          ]{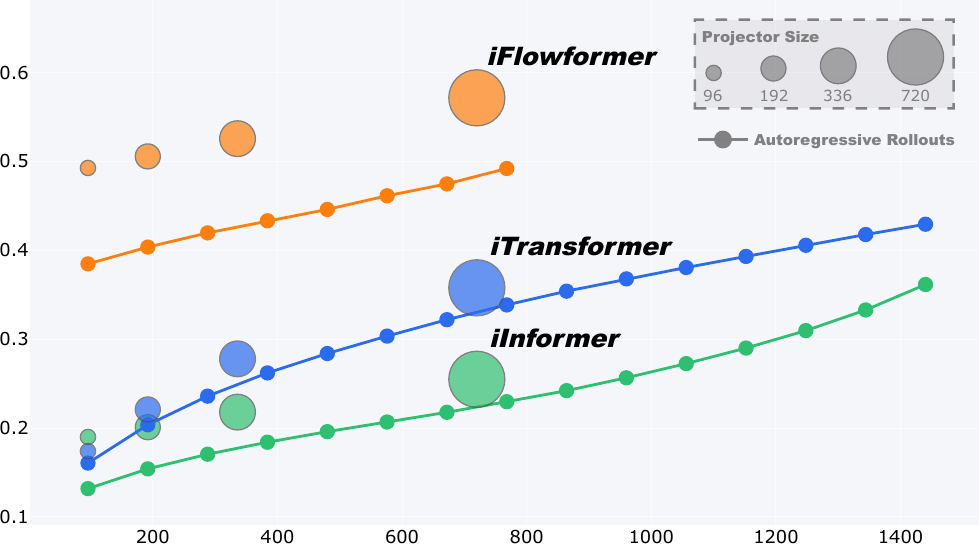}%
        };
        \node[rotate=90, anchor=south] at (img.west) {Mean Squared Error (MSE)};
        \node[anchor=north] at (img.south) {Prediction Length};
        \end{tikzpicture}
    }
    \caption{\textbf{Multiple forecasting models (vanilla training) vs. a single short-horizon forecasting model (our AROpt training)}. iTransformer (Weather), iInformer (Traffic), iFlowformer (Electricity). Our method outperforms vanilla training across varying horizons. It enables small models to produce flexible-length forecasts via AR rollouts without retraining and surpasses larger, specialized models on long-horizon forecasting.}
    \label{fig_plotly}
  \end{center}
\end{wrapfigure}

Transformer-based architectures have emerged as the dominant paradigm in time-series forecasting, having demonstrated excellent performance across diverse benchmarks. A key milestone was PatchTST \citep{nie2022time}, inspired by ViT \citep{dosovitskiy2020image}, which segments each univariate time series into subseries-level patches. Building on this, iTransformer \citep{liu2023itransformer} introduced an inverted design: instead of tokenizing along the temporal dimension, it treats variates as tokens. This approach has achieved state-of-the-art (SOTA) results in long-term forecasting, addressing limitations of prior temporal-token Transformers and outperforming them on many benchmarks.  

\begin{figure}[ht]
  \vskip 0.2in
  \begin{center}
    \centerline{\includegraphics[width=\columnwidth]{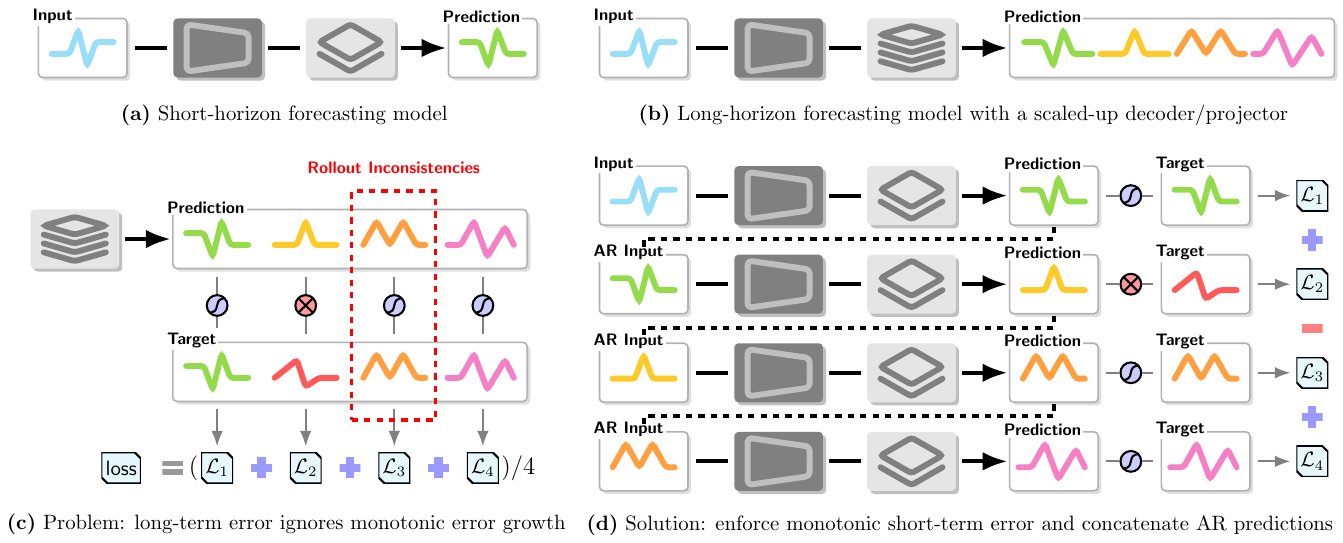}}
    \caption{
    \textbf{Direct long-term forecasting \textit{vs.} AR rollout forecasting}
      (a) A short-horizon forecasting model that predicts the next forecasting horizon using an encoder–projector architecture. This architecture is based solely on iTransformer, and the output window is strictly separated from the input window. In contrast, for Informer, Flashformer, and Flowformer, the projector is contained within the decoder, and they employ overlapping input–output windows. (b) A long-horizon forecasting model that reuses the same encoder but scales up the decoder/projector to directly predict longer future horizons. (c) Problem: direct long-horizon forecasting can lead to rollout inconsistency -- the model \textbf{forecasts a correct prediction patch after an incorrect one} or \textbf{reduces MSE as the forecasting horizon increases}. (d) We propose a new loss function that enforces monotonic error growth: short-term AR rollout predictions should exhibit increasing error $\mathcal{L}_t$ as the time step $t$ increases. Violations of this constraint indicate rollout inconsistency and are penalized in the total loss. These short-term AR rollout predictions are then concatenated to form long-horizon predictions.
    }
    \label{fig_pipeline}
  \end{center}
\end{figure}

Current time-series forecasting models typically address different forecasting horizons by scaling model capacity or training horizon-specific models \cite{nie2022time,wu2022timesnet,zhou2021informer,huang2022flowformer,dao2022flashattention,liu2023itransformer}. Such designs limit the flexibility of a single model to perform arbitrary-length forecasting. In contrast, causal language models (e.g., GPT \cite{radford2018improving}), naturally extend generation beyond the training output length through autoregressive (AR) rollout, where each prediction conditions on previously generated outputs.
Directly applying AR rollout to time-series forecasting remains challenging because prediction errors recursively propagate through subsequent inputs \cite{marcellino2006comparison,box2015time}.
Small prediction errors at early steps propagate and accumulate during long-horizon AR rollout, leading to rapid degradation in forecast quality \cite{marcellino2006comparison}. 
Consequently, inaccuracies introduced at early rollout steps accumulate over time and typically result in progressively larger forecasting errors. We emphasize that this error-growth phenomenon arises from the AR rollout process itself rather than from an intrinsic property of the underlying dynamical system. Even when the target dynamics exhibit seasonality or mean reversion, recursively feeding predicted values into future inputs generally causes prediction uncertainty to accumulate during rollout. This distinction motivates our optimization strategy.

In this work, we propose \textbf{AROpt}, a general optimization framework for supervised time-series forecasting. As illustrated in Fig. Fig.~\ref{fig_pipeline}(a \& b),existing forecasting models typically increase model size to support longer prediction horizons. Instead, AROpt explicitly optimizes the AR rollout process by collecting rollout errors and introducing a soft monotonic rollout regularizer into the training objective. The proposed regularizer encourages consistent error progression during rollout while preserving the original optimization target. Specifically, AROpt prioritizes reducing earlier rollout errors rather than encouraging larger errors at later horizons. The resulting short-term predictions can then be concatenated to generate flexible-length forecasts without modifying the model architecture.

Building upon this framework, our contributions are summarized as follows: (1) We propose a rollout-aware optimization framework that regularizes autoregressive error propagation through a soft monotonic rollout objective. The proposed objective preserves the original optimization target while encouraging stable AR rollout behavior. (2) We demonstrate that AROpt enables a single fixed forecasting model to generalize across short-, long-, and flexible-length prediction scenarios through AR rollout, reducing the need for horizon-specific model scaling. (3) Extensive experiments across multiple benchmark datasets and forecasting architectures show consistent improvements over strong supervised forecasting baselines. In addition, AROpt enables short-horizon forecasting models to generate accurate long-horizon forecasts through AR rollout while maintaining competitive computational efficiency.

\section{Related Work}
\label{sec:RelatedWork}

\subsection{Monotonic Error Growth in Forecasting Models}
Current high-performance sequential models for time series forecasting are typically based on the Transformer architecture. 
Traditionally, these forecasters employ an encoder-decoder structure as their core architecture with the same input/output data format as in standard Transformer training, where the input sequence and target sequence share an overlapping subsequence \cite{wu2021autoformer,nie2022time,zhou2022fedformer} (see Appendix~\ref{appendix:Extra Experiments}). 
In contrast, iTransformer introduces a major conceptual shift and achieves significantly stronger baseline performance \cite{liu2023itransformer}: its predictions strictly begin at  the time location after the input sequence (with no overlapping window). 
This design raises confusion, as it abandons the use of overlapping windows, which are beneficial for AR rollout consistency. In this work, we focus on the autoregressive rollout process rather than the underlying dynamics of the target time series. We treat monotonic error growth as a coarse empirical property of autoregressive error propagation that is consistently observed across benchmark datasets and architectures.

\subsection{Autoregressive Rollout}
\label{sec:AutoregressiveRollout}
Autoregressive rollout is a model inference pipeline that is widely used in language models \cite{vaswani2017attention,radford2019language}. 
By using the model predictions as subsequent inputs, a model can iteratively generate sequences of flexible length.
However, this approach does not work in time series forecasting due to two issues: (1) outputs are continuous values, in contrast to the inherent discrete nature of causal language modeling \cite{zhou2021informer, wu2021autoformer}, and (2) AR prediction errors increase with the forecasting horizon. Even a small error can alter the future trajectory, as errors accumulate rapidly \cite{marcellino2006comparison}. Inspired by the accept/reject mechanism from DeepSeek-V3 pre-training, we propose a rollout-aware optimization framework that improves the stability of autoregressive forecasting.  AROpt regularizes the temporal structure of AR rollout errors. Importantly, the proposed objective does not modify the optimization target; instead, it reweights the optimization emphasis across rollout steps to encourage more consistent error propagation during inference.

\subsection{Multi-Token Prediction}
The raw multi-token prediction serves as an auxiliary objective in language models, duplicating the model head to predict multiple future tokens in a single forward pass \cite{gloeckle2024better}. However, this approach often shows no significant improvement in model performance. In the DeepSeek-V3 technical report, the authors reformulate this concept by using a shared-head structure to prevent excessive parameter scaling, then incorporate the causal autoregressive dependency structure into multi-token prediction to selectively accept or reject the additional outputs \cite{liu2024deepseek}. In Section \ref{sec:algorithm}, we extend this accept/reject mechanism by integrating it with AR rollout. 
Instead, AROpt incorporates rollout-aware optimization directly into the training objective, making the method architecture-agnostic and applicable to existing forecasting backbones without introducing additional model parameters.

\subsection{Zero-shot Generation}
Zero-shot generation enables models to generalize to new tasks without requiring task-specific retraining or fine-tuning \cite{sanh2021multitask}. 
While this capability has been popularized through in-context learning in language and generative vision models \cite{radford2021learning,tian2024visual}, its potential in time-series forecasting remains a critical research area.
Traditional forecasting models typically require the target sequences to match the length of the model projector and predicted sequences. 
This rigidity imposes significant bottlenecks: longer prediction horizons place additional constraints on small datasets and increase computational costs during training. 
In contrast, models optimized for zero-shot generation overcome this restriction. For example, a model trained on short prediction horizons (e.g., 96 steps) can reliably make long-horizon forecasting (e.g., 720 steps) via AR rollout at inference time. 

Unlike recent time-series foundation models \cite{ansari2024chronos,woo2024unified,das2023decoder,liu2024timer} that rely on large-scale pretraining followed by zero-shot or fine-tuning evaluation, our work studies AR rollout under the fully supervised forecasting setting. AROpt is therefore complementary to foundation-model approaches, providing a general optimization strategy that can improve supervised forecasting backbones without requiring architectural modifications or large-scale pretraining.

\section{Algorithm}
\label{sec:algorithm}

\begin{algorithm}[t]
\caption{AR rollout objective function and concatenating short-term prediction for long-horizon forecasting.
Given historical context length $S$, rollout stride $T$, input--output overlap length $L$ (for iTransformer-style architectures, $L=0$), and number of rollout steps $n$. Let $\gamma \in (0,1)$ be the decay factor (default: $0.5$) and $\beta \in (0,0.5)$ the smoothing weight (default: $0.1$). $f(\cdot;\theta)$ denotes the inference function and $\mathcal{L}$ the loss function (default: MSE). $\mathrm{sg}(\cdot)$ denotes the stop-gradient operator.}
\label{alg:autoregressive_criterion}
\begin{algorithmic}[1]
\STATE {\bfseries Input:} Historical context and future ground-truth sequence $\boldsymbol{X} = \{x_0, \cdots, x_{S+nT-1}\}$
\STATE $\hat{\boldsymbol{X}}_{:S} \gets \boldsymbol{X}_{:S}$ (Pad the prediction sequence)
\STATE $\hat{\boldsymbol{X}}_{S:S+T} \gets f(\boldsymbol{X}_{:S}; \boldsymbol{\theta})_{L:L+T}$ (Generate the first prediction patch for AR rollout)
\STATE $e_1 \gets \mathcal{L}(\hat{\boldsymbol{X}}_{S:S+T}, \boldsymbol{X}_{S:S+T})$ (Compute the loss of the first prediction patch)
\STATE $\ell \gets e_1$
\FOR{$k \gets 1$ {\bfseries to} $n-1$}
    \STATE $\hat{\boldsymbol{X}}_{S+kT:S+(k+1)T} \gets f(\hat{\boldsymbol{X}}_{kT:S+kT}; \boldsymbol{\theta})_{L:L+T}$ (Generate the $(k+1)$-th prediction patch)
    \STATE $e_{k+1} \gets \mathcal{L}(\hat{\boldsymbol{X}}_{S+kT:S+(k+1)T}, \boldsymbol{X}_{S+kT:S+(k+1)T})$ (Compute the loss of the $(k+1)$-th patch)
    \STATE $\ell \gets \ell +\gamma^k ((1-\beta)e_{k+1}+\beta|e_{k+1}-\text{sg}(e_k)|)$ (Accumulate the discounted loss)
\ENDFOR
\STATE {\bfseries Return:} total loss $\ell$ and the long-term prediction \(\hat{\boldsymbol{X}}_{S-L:S+nT} = \{ \hat{x}_{S-L}, \cdots, \hat{x}_{S+nT-1} \}\)
\end{algorithmic}
\end{algorithm}

The pseudocode of our proposed method is presented in \ref{alg:autoregressive_criterion}.  Let \( f(\cdot; \theta): \mathbb{R}^S \mapsto \mathbb{R}^{L+T} \) denote the model inference function with parameters \( \theta \), where the input is a sequence with length \( S \), the target has length \( L+T \), and \( L \) is the overlap length. We are interested in extending this mapping to \( \mathbb{R}^{S} \mapsto \mathbb{R}^{L+nT} \) via $n$-step AR rollout. This extension aligns the model with the following input-output or input-target structure:

\[
\includegraphics[width=0.8\columnwidth]{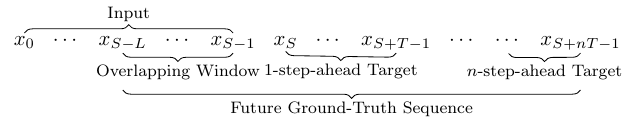}
\]

Let \( \{ \hat{x}_{S-L}, \dots, \hat{x}_{S+nT-1} \} \) denote the final expected long-term predictions, initialized as:

\begin{equation}
    (\hat{x}_{S-L}, \dots, \hat{x}_{S+T-1}) = f(x_0, \cdots, x_{S-1};\theta).
    \label{eqn:initial}
\end{equation}

We then use the last $S$ elements from the $k$-step prediction patch as the input to perform the ($k+1$)-step prediction patch:

\begin{equation}
    (\hat{x}_{S+kT-L}, \dots, \hat{x}_{S+(k+1)T-1}) = f(\hat{x}_{kT}, \dots, \hat{x}_{S+kT-1};\theta).
    \label{eqn:rollout}
\end{equation}

Subsequently, we repeat this AR rollout step until the prediction sequence of length \(L + nT\) is fully filled.

In AR rollout forecasting, prediction errors generally accumulate as the rollout horizon increases because each prediction recursively depends on previously generated predictions. For example, in traffic forecasting (see Table \ref{tab:ar}), all models using four days of input (length 96) achieve lower MSE for short-term prediction (length 96) than for long-term prediction (30 days, length 720). We emphasize that this monotonic error-growth assumption characterizes the AR rollout process rather than the underlying dynamics of the target time series. Although forecasting errors may fluctuate locally due to stochasticity, periodicity, seasonality, or transient stabilization effects, we treat monotonic error growth as a coarse empirical tendency commonly observed across benchmark datasets and forecasting architectures. Consequently, when later rollout predictions become substantially more accurate than earlier rollout predictions, this may indicate inconsistency between training and inference rollout behavior. To discourage such severe rollout inconsistencies, we incorporate a soft monotonic error-growth regularization term into the minimization of the MSE loss:

\begin{subequations}
\begin{align}
\min_{\theta} \quad &
\frac{1}{n}\sum_{t=S}^{S+nT-1} |x_t-\hat{x}_t|^2 \\
\text{subject to} \quad &
|x_t-\hat{x}_t| \ge |x_{t-1}-\hat{x}_{t-1}|,\ \forall t. \label{eq:condition}
\end{align}
\end{subequations}

Because the mini-batch training strategy renders the MSE loss a stochastic scalar function, traditional constrained optimization methods, such as the Lagrange multiplier method and barrier methods \cite{boyd2004convex}, are not applicable here. 
To address this issue, we introduce a rollout regularization term that encourages the model to generate outputs satisfying the \textbf{temporal monotonicity condition} (Eq.~\ref{eq:condition}) during training.
At the $k$-th step of the AR rollout, the MSE is taken over the entire prediction patch:

\[
e_k = \frac{1}{T} \sum_{t=S+kT}^{S+(k+1)T-1} | x_t - \hat{x}_t |^2.
\]

Rather than enforcing (Eq.~\ref{eq:condition}) as a hard constraint, we introduce the following rollout reward that softly encourages the temporal monotonicity condition while preserving the original MSE optimization objective.

\begin{equation}
r_k :=  
\begin{cases} 
e_0 &\ \text{if~} k=0, \\
-(1-\beta) e_k - \beta | e_k - \text{sg}(e_{k-1}) | &\ \text{if~} k > 1.
\end{cases}
\label{eqn:penalty}
\end{equation}

The hyperparameter $  \beta \in (0, 0.5)  $ controls the weight of the penalty term, while $  \text{sg}(\cdot)  $ denotes the stop-gradient operator, which acts as the identity function during the forward pass but prevents gradient computation through backpropagation \cite{chen2021exploring}. In this case, the gradient norm of the reward decreases upon detecting rollout inconsistencies:
\begin{equation*}
\nabla_{\theta} r_k =  
\begin{cases}
-\nabla_{\theta} e_k & \text{if~} e_k > e_{k-1}, \\
-(1-\beta)\nabla_{\theta} e_k & \text{if~} e_k = e_{k-1}, \\
-(1-2\beta)\nabla_{\theta} e_k & \text{if~} e_k < e_{k-1}.
\end{cases}
\end{equation*}
The above equation illustrates the optimization emphasis of AROpt. When rollout inconsistency is detected (i.e., later rollout errors become unexpectedly smaller than earlier ones), the corresponding gradient magnitude is reduced. Conversely, when the rollout error follows the expected accumulation trend, larger gradients are preserved. Importantly, this mechanism changes only the optimization emphasis across rollout steps rather than the optimization target itself, since every rollout error remains positively weighted for $0 < \beta < 0.5$.

Finally, we define the objective as a discounted weighted sum of rollout losses with discount factor $\gamma$:

\begin{equation}
\ell := - \sum_{k=0}^{n-1} \gamma^k r_k.
\label{eqn:loss}
\end{equation}

Although Eq. (\ref{eqn:loss}) is expressed in a reward-style discounted form, it is not a reinforcement learning objective. The reward notation is introduced solely to describe temporally weighted rollout optimization. The resulting objective is fully differentiable and can be optimized directly using standard backpropagation without policy-gradient estimation or any modification to the training algorithm.
In this work, \( \gamma \) is fixed to be 0.5. The reason is that when the per-step errors $e_k$ increase monotonically but remain of comparable magnitude, the effective accumulated loss is given by

\begin{equation}
\ell = \frac{1-\gamma^{n}}{1-\gamma} \mathcal{O} (e_0) \approx 2 \mathcal{O} (e_0),
\label{eq:estimation}
\end{equation}
for large $n$. Thus, the objective has roughly twice the magnitude of a conventional MSE in short-term prediction. Empirically, we find that $\beta = 0.1$ yields substantial performance improvements across diverse baseline models. This choice is based on our grid search experiments. We also provide the corresponding hyperparameter sensitivity analysis in Appendix \ref{appendix:Sensitivity Analysis}.

\section{Convergence Analysis}
\label{sec:ConvergenceAnalysis}
The following analysis is intended to provide qualitative intuition regarding optimization stability rather than a formal convergence proof for non-convex Transformer training. Since the proposed objective is a positively weighted linear combination of per-step MSE losses, our goal is to show that it preserves bounded gradient behavior under assumptions commonly adopted in stochastic optimization.

From definitions (\ref{eqn:penalty}) and (\ref{eqn:loss}), the objective can be expressed as a positively weighted linear combination of the per-step MSE losses (see Appendix~\ref{appendix:Asymptotic and Non-asymptotic Analysis}). Consequently, \(\nabla \ell\) is also a linear combination of the corresponding MSE gradients, indicating that AROpt preserves stable optimization behavior during model training. Assume the error $e_k$ of the $k$-step prediction patch has bounded gradient norm \( \mathbb{E}[\|\nabla_\theta e_k\|] \le d  \), we define the centered noise for the gradient of reward function:

\[
G_k := \nabla_\theta r_k - \mathbb{E}[\nabla_\theta r_k].
\]

Then the centered gradient estimator is 

\begin{equation}
\nabla_\theta \ell - \mathbb{E}[\nabla_\theta \ell] = -\sum_{k=0}^{n-1} \gamma^k G_k.
\label{eqn:ca1}
\end{equation}

By the Cauchy–Schwarz inequality,

\begin{equation}
\mathbb{E}\left[ \left\| \sum_{k=0}^{n-1} \gamma^k G_k \right\|^2 \right] \le \left( \sum_{k=0}^{n-1} \gamma^k \sqrt{ \mathbb{E} [ \|G_k\|^2 ] } \right)^2.
\label{eqn:ca2}
\end{equation}

From the definition (\ref{eqn:penalty}),

\[
\| \nabla_\theta r_k \| \le \| \nabla_\theta e_k \|
\]
Thus,

\begin{equation}
\mathbb{E} [ \|G_k\|^2 ] \le \mathbb{E} [ \|\nabla_\theta r_k\|^2 ] \le \mathbb{E} [ \|\nabla_\theta e_k\|^2 ] \le d^2.
\label{eqn:ca3}
\end{equation}

Substituting the inequalities (\ref{eqn:ca1}), (\ref{eqn:ca2}), and (\ref{eqn:ca3}), we obtain the bound of the gradient norm

\[
\mathbb{E} \left[ \| \nabla_\theta \ell \|^2 \right] = \mathbb{E}\left[ \left\| \sum_{k=0}^{n-1} \gamma^k G_k \right\|^2 \right] \le \left( \sum_{k=0}^{n-1} \gamma^k \, d \right)^2 < 4 d^2.    
\]

This estimation suggests that AROpt preserves the same bounded-gradient property as the standard MSE objective under the above assumptions, indicating comparable optimization stability during training. We also provide an extra detailed analysis in Appendix~\ref{appendix:Asymptotic and Non-asymptotic Analysis}.

\section{Experiments}
\label{sec:Experiments}

Our proposed approach is model-agnostic and integrates seamlessly with SOTA models (iTransformer and its variants) without requiring architectural changes. 
Specifically, its optimization method operates on a novel loss function and data pipeline to enable AR rollouts for varying-length predictions across multivariate time-series forecasting benchmarks.  Experiments demonstrate that our method not only enhances model performance for short-term forecasting, but also enables short-horizon forecasting models to generate high-accuracy long-term predictions via AR rollouts, often outperforming models specifically scaled for longer-term forecasting. 

\subsection{Setup}

We adopt the widely used benchmarks from the iTransformer study \cite{liu2023itransformer}, including Exchange, Electricity (ECL), Traffic, Weather, Solar-Energy, PEMS, and ETTh.  These datasets capture diverse real-world scenarios: Exchange consists of daily exchange rates for eight countries; Electricity records hourly power consumption across 321 clients; Traffic measures hourly road occupancy from 862 San Francisco Bay area highway sensors (2015–2016); Weather tracks 21 meteorological variables at 10-minute intervals in Germany (2020); Solar-Energy monitors the solar power production from137 PV plants in 2006; and PEMS datasets reflect traffic flow across the California state highway system. 

Since iTransformer and its variants already achieve SOTA results on standard benchmarks, we primarily apply our proposed method to four inverted Transformer architectures: iTransformer, iInformer, iFlowformer, and iFlashformer. Baseline numbers are taken directly from that work, where each model is trained independently for every prediction length. For ablation studies, we adopt the same Adam optimizer and relevant configurations as in that work; experiments run on a single NVIDIA RTX A6000 GPU.

Our approach introduces three additional hyperparameters: the reward discount factor $\gamma$, the penalty weight $\beta$, and the number of AR rollout steps $n$. We intentionally fix these hyperparameters (see the default settings in Algorithm \ref{alg:autoregressive_criterion}) across all experiments to demonstrate that the proposed method does not rely on extensive hyperparameter tuning to achieve performance gains. Model configurations and random seed follow the iTransformer GitHub repository.

\subsection{Main Results and Analysis}
\label{sec:Results}
Table~\ref{tab:ar} reports detailed forecasting performance across diverse datasets, prediction length, and model variants. Blue and red values indicate the best MSE and MAE for each task–model pair, respectively, and bold-underlined values mark the overall best result across all models for each task. 
\begin{wrapfigure}[20]{r}{0.5\textwidth}
  \vspace{-20pt}
  \vskip 0.2in
  \begin{center}
    \centerline{\includegraphics[width=0.5\columnwidth]{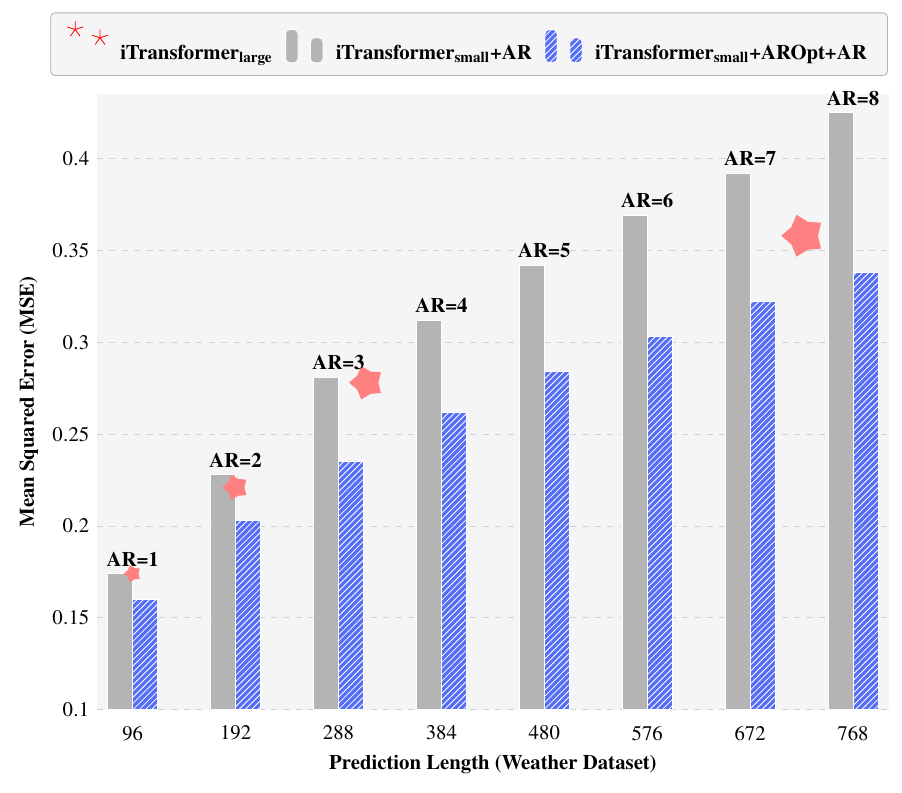}}
    \caption{
    \textbf{Ablation study.} Red stars: larger models with different output lengths. Gray bars: a small model with AR rollout. Blue bars: the same small model with AROpt and AR rollout.
    }
    \label{fig_ablation_study}
  \end{center}
\end{wrapfigure}
Our approach ranks first in the majority of cases and outperforms baselines. While prior work \cite{wu2021autoformer,zhou2021informer, wu2022timesnet, dao2022flashattention, huang2022flowformer,nie2022time,liu2023itransformer} emphasizes scaling up the model projector is required for long-term forecasting, our approach instead enables a single model to generate predictions of flexible length by concatenating AR rollout outputs. We denote this procedure as ``(AR=$k$)" to indicate $k$-steps AR rollouts.

\begin{table*}[t!]
\centering
\caption{Forecasting performance on the Electricity, Traffic, and Weather datasets with fixed lookback length $S = 96$ and prediction lengths $T \in \{96, 192, 336, 720\}$. 
Task names are formatted by concatenating the dataset name + prediction length (e.g., ``ECL\_96", ``Traffic\_720"). Baseline results are taken from the iTransformer paper. 
Prediction length under our method is extended to $(\operatorname{AR} \times T)$.
}
\label{tab:ar}
\begin{tabular}{l|lr|ll|ll|ll|ll}
\toprule
& &
& \multicolumn{2}{c|}{iTransformer} 
& \multicolumn{2}{c|}{iInformer} 
& \multicolumn{2}{c|}{iFlowformer} 
& \multicolumn{2}{c}{iFlashformer} \\
Task & \multicolumn{2}{c|}{Pred.\ Len.} & MSE & MAE & MSE & MAE & MSE & MAE & MSE & MAE \\
\midrule

\multirow{5}{*}{ECL\_96} & 96 & Baseline 
& 0.148 & 0.240 & 0.190 & 0.286 & 0.183 & 0.267 & 0.178 & 0.265 \\ 
& 96 & (AR=1)  
& {\color{blue}0.141} & {\color{red}0.239} & \underline{\textbf{\color{blue}0.131}} & \underline{\textbf{\color{red}0.232}} & {\color{blue}0.141} & {\color{red}0.241} & {\color{blue}0.142} & {\color{red}0.241} \\
& 192 & (AR=2)  
& 0.163 & 0.260 & 0.154 & 0.254 & 0.167 & 0.266 & 0.169 & 0.267 \\
& 384 & (AR=4)    
& 0.194 & 0.288 & 0.184 & 0.282 & 0.207 & 0.301 & 0.206 & 0.300 \\
& 768 & (AR=8)    
& 0.238 & 0.324 & 0.229 & 0.320 & 0.263 & 0.345 & 0.257 & 0.340 \\ \midrule

\multirow{4}{*}{ECL\_192} & 192 & Baseline 
& 0.162 & \underline{\textbf{\color{red}0.253}} & 0.201 & 0.297 & 0.192 & 0.277 & 0.189 & 0.276 \\
& 192 & (AR=1)
& {\color{blue}0.161} & 0.258 & {\color{blue}0.161} & {\color{red}0.262} & \underline{\textbf{\color{blue}0.158}} & {\color{red}0.256} & {\color{blue}0.162} & {\color{red}0.261} \\
& 384 & (AR=2)  
& 0.189 & 0.285 & 0.191 & 0.290 & 0.186 & 0.282 & 0.195 & 0.290 \\
& 768 & (AR=4)  
& 0.230 & 0.318 & 0.230 & 0.322 & 0.223 & 0.312 & 0.239 & 0.325 \\ \midrule

\multirow{3}{*}{ECL\_336} & 336 & Baseline 
& 0.178 & \underline{\textbf{\color{red}0.269}} & 0.218 & 0.315 & 0.210 & 0.295 & 0.207 & 0.294 \\
& 336 & (AR=1)
& \underline{\textbf{\color{blue}0.177}} & 0.275 & {\color{blue}0.187} & {\color{red}0.288} & {\color{blue}0.178} & {\color{red}0.276} & {\color{blue}0.184} & {\color{red}0.282} \\
& 672 & (AR=2)
& 0.209 & 0.302 & 0.230 & 0.323 & 0.213 & 0.306 & 0.221 & 0.312 \\ \midrule

\multirow{2}{*}{ECL\_720} & 720 & Baseline 
& 0.225 & 0.317 & 0.255 & 0.347 & 0.255 & 0.332 & 0.251 & 0.329 \\
& 720 & (AR=1)
& {\color{blue}0.207} & {\color{red}0.300} & \underline{\textbf{\color{blue}0.203}} & \underline{\textbf{\color{red}0.296}} & {\color{blue}0.210} & {\color{red}0.303} & {\color{blue}0.222} & {\color{red}0.314} \\ \midrule \midrule

\multirow{5}{*}{Traffic\_96} & 96 & Baseline 
& {\color{blue}0.395} & 0.268 & 0.632 & 0.367 & 0.493 & 0.339 & 0.464 & 0.320 \\ 
& 96 & (AR=1)  
& 0.399 & {\color{red}0.267} & {\color{blue}0.394} & {\color{red}0.262} & \underline{\textbf{\color{blue}0.384}} & {\color{red}0.259} & {\color{blue}0.385} & \underline{\textbf{\color{red}0.256}} \\
& 192 & (AR=2)  
& 0.414 & 0.277 & 0.412 & 0.274 & 0.403 & 0.271 & 0.398 & 0.267 \\
& 384 & (AR=4)  
& 0.440 & 0.298 & 0.442 & 0.297 & 0.433 & 0.293 & 0.419 & 0.283 \\
& 768 & (AR=8)  
& 0.496 & 0.338 & 0.502 & 0.343 & 0.475 & 0.323 & 0.448 & 0.303 \\ \midrule

\multirow{4}{*}{Traffic\_192} & 192 & Baseline 
& 0.417 & 0.276 & 0.641 & 0.370 & 0.506 & 0.345 & 0.479 & 0.326 \\
& 192 & (AR=1)  
& {\color{blue}0.413} & {\color{red}0.274} & {\color{blue}0.408} & {\color{red}0.269} & {\color{red}0.402} & {\color{blue}0.269} & \underline{\textbf{\color{blue}0.393}} & \underline{\textbf{\color{red}0.266}} \\
& 384 & (AR=2)  
& 0.436 & 0.290 & 0.433 & 0.285 & 0.426 & 0.285 & 0.411 & 0.278 \\
& 768 & (AR=4)  
& 0.479 & 0.319 & 0.478 & 0.316 & 0.471 & 0.314 & 0.441 & 0.297 \\ \midrule

\multirow{3}{*}{Traffic\_336} & 336 & Baseline 
& 0.433 & 0.283 & 0.663 & 0.379 & 0.526 & 0.355 & 0.501 & 0.337 \\
& 336 & (AR=1)  
& {\color{blue}0.427} & {\color{red}0.282} & {\color{blue}0.424} & {\color{red}0.278} & {\color{blue}0.418} & {\color{red}0.277} & \underline{\textbf{\color{blue}0.396}} & \underline{\textbf{\color{red}0.270}} \\
& 672 & (AR=2)  
& 0.463 & 0.304 & 0.462 & 0.302 & 0.456 & 0.300 & 0.421 & 0.285 \\ \midrule

\multirow{2}{*}{Traffic\_720} & 720 & Baseline 
& 0.467 & {\color{red}0.302} & 0.713 & 0.405 & 0.572 & 0.381 & 0.524 & 0.350 \\
& 720 & (AR=1)  
& {\color{blue}0.466} & 0.305 & {\color{blue}0.465} & {\color{red}0.300} & {\color{blue}0.458} & {\color{red}0.300} & \underline{\textbf{\color{blue}0.421}} & \underline{\textbf{\color{red}0.285}} \\ \midrule \midrule

\multirow{5}{*}{Weather\_96} & 96 & Baseline 
& 0.174 & 0.214 & 0.180 & 0.251 & 0.183 & 0.223 & 0.177 & 0.218 \\ 
& 96 & (AR=1)  
& {\color{blue}0.160} & \underline{\textbf{\color{red}0.210}} & \underline{\textbf{\color{blue}0.147}} & {\color{red}0.201} & {\color{blue}0.161} & {\color{red}0.211} & {\color{blue}0.162} & {\color{red}0.212} \\
& 192 & (AR=2)  
& 0.203 & 0.235 & 0.190 & 0.242 & 0.203 & 0.250 & 0.203 & 0.250 \\
& 384 & (AR=4)  
& 0.262 & 0.294 & 0.248 & 0.286 & 0.259 & 0.292 & 0.260 & 0.293 \\
& 768 & (AR=8)  
& 0.338 & 0.342 & 0.322 & 0.333 & 0.333 & 0.339 & 0.333 & 0.310 \\ \midrule

\multirow{4}{*}{Weather\_192} & 192 & Baseline 
& 0.221 & 0.254 & 0.244 & 0.318 & 0.231 & 0.262 & 0.229 & 0.261 \\ 
& 192 & (AR=1)  
& {\color{blue}0.204} & {\color{red}0.251} & \underline{\textbf{\color{blue}0.194}} & \underline{\textbf{\color{red}0.246}} & {\color{blue}0.204} & {\color{red}0.251} & {\color{blue}0.202} & {\color{red}0.248} \\
& 384 & (AR=2)  
& 0.261 & 0.295 & 0.251 & 0.290 & 0.261 & 0.294 & 0.259 & 0.292 \\
& 768 & (AR=4)  
& 0.335 & 0.342 & 0.326 & 0.337 & 0.333 & 0.340 & 0.334 & 0.341 \\ \midrule

\multirow{3}{*}{Weather\_336} & 336 & Baseline 
& 0.278 & 0.296 & 0.282 & 0.343 & 0.286 & 0.301 & 0.283 & 0.300 \\ 
& 336 & (AR=1)  
& {\color{blue}0.251} & {\color{red}0.286} & \underline{\textbf{\color{blue}0.246}} & {\color{red}0.286} & {\color{blue}0.249} & \underline{\textbf{\color{red}0.285}} & {\color{blue}0.250} & {\color{red}0.287} \\
& 672 & (AR=2)  
& 0.320 & 0.331 & 0.318 & 0.332 & 0.317 & 0.330 & 0.319 & 0.332 \\ \midrule

\multirow{2}{*}{Weather\_720} & 720 & Baseline 
& 0.358 & 0.349 & 0.377 & 0.409 & 0.363 & 0.352 & 0.359 & \underline{\textbf{\color{red}0.251}} \\ 
& 720 & (AR=1)  
& {\color{blue}0.329} & {\color{red}0.337} & {\color{blue}0.326} & {\color{red}0.338} & \underline{\textbf{\color{blue}0.325}} & {\color{red}0.335} & {\color{blue}0.328} & 0.337 \\

\bottomrule
\end{tabular}
\end{table*}

\paragraph{Short-term Forecasting.} The baseline models are trained using a conventional training process, while our models are trained with AROpt. When the AR rollout step is set to 1, AR rollout is disabled during inference. Under this setting, with the prediction length set to 96 (matching the input length), the model forecasts predictions over a standard short horizon. Table \ref{tab:ar} reports the consistent performance gains of the models in short-term forecasting. For example, on Electricity (prediction length 96), iInformer MSE drops from 0.190 to 0.131. On Traffic, iFlashformer reduces MSE from 0.464 to 0.385, and on Weather, iFlowformer reduces MSE from 0.183 to 0.161.

\paragraph{Long-term Forecasting.} When the AR rollout step is set to 1 and the prediction length is set to 720 (much longer than the input length of 96), the model generates forecasts over a standard long horizon. Table \ref{tab:ar} reports consistent performance gains of the models in long-term forecasting. For example, on Electricity (prediction length 720), iInformer MSE drops from 0.255 to 0.203. On Traffic, iFlashformer reduces MSE from 0.524 to 0.421, and on Weather, iFlowformer reduces MSE from 0.363 to 0.325.

\paragraph{Flexible-length Forecasting} When the AR rollout step is not equal to 1, the models operate in an autoregressive rollout mode, generating prediction patches that are concatenated to form longer predictions. Our empirical results demonstrate that smaller models trained using AROpt outperform larger models. For example, on Electricity, the small iInformer (0.229, length \(96 \times 8=768\)) outperforms the large iInformer (0.255, length 720). On Traffic, the small iFlowformer (0.572, length \(96 \times 8=768\)) outperforms the large iFlowformer (0.475, length 720). On Weather, the small iTransformer (0.338, length \(96 \times 8=768\)) outperforms 0.358 (length 720). We note that AR rollout introduces additional inference iterations compared to direct one-shot prediction. Our focus in this work is forecasting quality and flexible horizon generalization rather than inference latency optimization. In many practical settings, the ability to reuse a single short-horizon model across multiple forecasting lengths may nevertheless reduce overall deployment and retraining costs.

So far, our empirical results demonstrate the following performance ranking:
\[
\text{model}_{\text{large}}+\text{AROpt}
> \text{model}_{\text{small}}+\text{AROpt}+\text{AR Rollout}
> \text{model}_{\text{large}}
> \text{model}_{\text{small}}+\text{AR Rollout},
\]

which highlights the contribution of our AROpt. This ranking can be easily understood from the visualization in Fig.~\ref{fig_ablation_study}.

To further support our conclusions, Appendix~\ref{appendix:Extra Experiments} presents experiments on additional models and datasets to assess robustness. Appendix~\ref{appendix:Ablation Study} provides detailed ablation studies, evaluating model performance under the standard training method and AR rollout inference mode.

\subsection{SOTA Model Ranking Reversal}
Under vanilla direct training, the base iTransformer consistently outperforms its variants iInformer, iFlowformer, and iFlashformer across datasets and horizons, establishing it as the strongest baseline among inverted Transformer models. In contrast, our proposed optimization strategy dramatically reshuffles this ordering. Previously underperforming variants now frequently achieve the lowest MSE and MAE values, often surpassing both the vanilla iTransformer and the AR-optimized base iTransformer.
This reversal is consistent across short-, medium-, and long-term forecasting tasks. On Traffic, for example, optimized iFlashformer sets new SOTA results at multiple horizons (e.g., MSE 0.385 at prediction length 96, 0.393 at prediction length 192), while optimized iInformer and iFlowformer lead on select Electricity and Weather subtasks. The base iTransformer remains competitive but no longer dominates. We additionally compare AROpt with recent time-series foundation models \cite{ansari2024chronos,woo2024unified,das2023decoder,liu2024timer} under zero-shot and full fine-tuning evaluation protocols; the detailed results and discussion are provided in Appendix \ref{app:Comparison with Time-Series Foundation Models}.




\section{Future Work}
\label{sec:FutureWork}

While the proposed optimization method enables flexible-length forecasting with a fixed model and improves performance across diverse datasets and architectures, several directions remain for further study.
First, although our experiments reveal an approximately monotonic relationship between error accumulation and prediction length (projector length $\times$ AR rollout steps), this behavior has so far been studied only empirically. A more rigorous analysis of rollout error propagation remains future work, including potential bounds on long-horizon MSE/MAE under realistic noise and bias amplification assumptions. 
The proposed rollout regularization is motivated by empirical observations on standard forecasting benchmarks. Its effectiveness for highly chaotic, strongly mean-reverting, or irregularly sampled systems remains an important direction for future investigation. In addition, extremely long autoregressive horizons may still suffer from cumulative distribution shift and error amplification.
Future work will explore more adaptive rollout regularization strategies and broader applications of AR-optimized forecasting models across diverse time-series domains.

\section{Conclusion}
\label{sec:Conclusion}

In this work, we introduce a novel loss function and training pipeline to address one of the most persistent challenges in modern time-series forecasting, optimizing AR rollout. Our method enables a single fixed model to generate high-quality long-term forecasts of flexible length through AR rollout without requiring architectural changes or retraining.
Extensive experiments across diverse datasets demonstrate that our method consistently improves the performance of iTransformer and its recent variants, achieving MSE reductions exceeding 10\% for both short- and long-term forecasting. Moreover, our proposed approach is preferable to directly training a long-horizon forecasting model, as it enables effective reuse of a short-horizon forecasting model for long-term forecasting, often outperforming specialized long-horizon forecasting models with larger-scale projectors on the evaluated benchmarks.
We believe this work opens promising avenues for optimization methods in autoregressive time-series forecasting.

\bibliographystyle{plainnat}
\bibliography{references}


\appendix

\section{Extra Experiments}
\label{appendix:Extra Experiments}

\begin{table*}[t!]
\centering
\caption{Forecasting performance on additional model architectures with fixed lookback length $S = 96$ and prediction lengths $T \in \{96, 192, 336, 720\}$. 
Task names are formatted by concatenating the dataset name + prediction length (e.g., ``ECL\_96", ``Traffic\_720"). Baseline results are taken from the iTransformer paper. 
Prediction length under our method is extended to $(\operatorname{AR} \times T)$.
}
\label{tab:sr}
\begin{tabular}{l|lr|ll|ll|ll|ll}
\toprule
& &
& \multicolumn{2}{c|}{Transformer} 
& \multicolumn{2}{c|}{Informer} 
& \multicolumn{2}{c|}{Flowformer} 
& \multicolumn{2}{c}{Flashformer} \\
Task & \multicolumn{2}{c|}{Pred.\ Len.} & MSE & MAE & MSE & MAE & MSE & MAE & MSE & MAE \\
\midrule

\multirow{5}{*}{ECL\_96} & 96 & Baseline 
& 0.260 & 0.358 & 0.274 & 0.368 & 0.215 & 0.320 & 0.259 & 0.357 \\ 
& 96 & (AR=1)  
& 0.235 & 0.344 & 0.272 & 0.375 & 0.242 & 0.342 & 0.246 & 0.345 \\
& 192 & (AR=2)  
& 0.251 & 0.356 & 0.276 & 0.377 & 0.251 & 0.348 & 0.255 & 0.351 \\ 
& 384 & (AR=4)    
& 0.263 & 0.364 & 0.279 & 0.378 & 0.261 & 0.351 & 0.261 & 0.356 \\ 
& 768 & (AR=8)    
& 0.266 & 0.366 & 0.285 & 0.385 & 0.266 & 0.354 & 0.260 & 0.355 \\  \midrule

\multirow{4}{*}{ECL\_192} & 192 & Baseline 
& 0.266 & 0.367 & 0.296 & 0.386 & 0.259 & 0.355 & 0.274 & 0.374 \\
& 192 & (AR=1)
& 0.264 & 0.360 & 0.292 & 0.386 & 0.248 & 0.346 & 0.261 & 0.355 \\
& 384 & (AR=2)  
& 0.272 & 0.364 & 0.299 & 0.391 & 0.250 & 0.347 & 0.275 & 0.365 \\
& 768 & (AR=4)  
& 0.284 & 0.371 & 0.309 & 0.398 & 0.249 & 0.347 & 0.287 & 0.371 \\ \midrule

\multirow{3}{*}{ECL\_336} & 336 & Baseline 
& 0.280 & 0.375 & 0.300 & 0.394 & 0.296 & 0.383 & 0.310 & 0.396 \\
& 336 & (AR=1)
& 0.285 & 0.369 & 0.326 & 0.409 & 0.273 & 0.362 & 0.283 & 0.366 \\
& 672 & (AR=2)
& 0.297 & 0.394 & 0.330 & 0.410 & 0.287 & 0.369 & 0.303 & 0.376 \\ \midrule

\multirow{2}{*}{ECL\_720} & 720 & Baseline 
& 0.302 & 0.386 & 0.373 & 0.439 & 0.296 & 0.380 & 0.298 & 0.383 \\
& 720 & (AR=1)
& 0.299 & 0.377 & 0.335 & 0.415 & 0.281 & 0.363 & 0.294 & 0.371 \\ \midrule \midrule

\multirow{5}{*}{Traffic\_96} & 96 & Baseline 
& 0.647 & 0.357 & 0.719 & 0.391 & 0.691 & 0.393 & 0.641 & 0.348 \\
& 96 & (AR=1)  
& 0.637 & 0.334 & 0.713 & 0.387 & 0.682 & 0.359 & 0.645 & 0.331 \\
& 192 & (AR=2)  
& 0.640 & 0.337 & 0.721 & 0.393 & 0.686 & 0.361 & 0.650 & 0.334 \\
& 384 & (AR=4)  
& 0.645 & 0.342 & 0.740 & 0.405 & 0.692 & 0.365 & 0.654 & 0.336 \\
& 768 & (AR=8)  
& 0.653 & 0.351 & 0.766 & 0.424 & 0.699 & 0.373 & 0.657 & 0.341 \\ \midrule

\multirow{4}{*}{Traffic\_192} & 192 & Baseline 
& 0.649 & 0.356 & 0.696 & 0.379 & 0.729 & 0.419 & 0.648 & 0.358 \\
& 192 & (AR=1)  
& 0.633 & 0.331 & 0.726 & 0.401 & 0.668 & 0.353 & 0.711 & 0.385 \\
& 384 & (AR=2)  
& 0.637 & 0.335 & 0.724 & 0.401 & 0.676 & 0.358 & 0.716 & 0.388 \\
& 768 & (AR=4)  
& 0.644 & 0.342 & 0.727 & 0.407 & 0.687 & 0.368 & 0.719 & 0.393 \\ \midrule

\multirow{3}{*}{Traffic\_336} & 336 & Baseline 
& 0.667 & 0.364 & 0.777 & 0.420 & 0.756 & 0.423 & 0.670 & 0.364 \\
& 336 & (AR=1)  
& 0.694 & 0.377 & 0.726 & 0.398 & 0.698 & 0.381 & 0.729 & 0.386 \\
& 672 & (AR=2)
& 0.709 & 0.389 & 0.719 & 0.397 & 0.711 & 0.391 & 0.739 & 0.394 \\ \midrule

\multirow{2}{*}{Traffic\_720} & 720 & Baseline 
& 0.697 & 0.376 & 0.864 & 0.472 & 0.825 & 0.449 & 0.673 & 0.354 \\
& 720 & (AR=1)  
& 0.657 & 0.356 & 0.800 & 0.454 & 0.740 & 0.412 & 0.640 & 0.580 \\ \midrule \midrule

\multirow{5}{*}{Weather\_96} & 96 & Baseline 
& 0.395 & 0.427 & 0.300 & 0.384 & 0.182 & 0.233 & 0.388 & 0.425 \\
& 96 & (AR=1)  
& 0.359 & 0.401 & 0.331 & 0.395 & 0.505 & 0.481 & 0.493 & 0.470 \\
& 192 & (AR=2)  
& 0.366 & 0.401 & 0.351 & 0.410 & 0.548 & 0.492 & 0.518 & 0.480 \\
& 384 & (AR=4)  
& 0.405 & 0.420 & 0.365 & 0.420 & 0.596 & 0.509 & 0.549 & 0.492 \\
& 768 & (AR=8)  
& 0.445 & 0.439 & 0.377 & 0.427 & 0.626 & 0.520 & 0.580 & 0.506 \\ \midrule

\multirow{4}{*}{Weather\_192} & 192 & Baseline 
& 0.619 & 0.560 & 0.598 & 0.544 & 0.250 & 0.288 & 0.619 & 0.560 \\
& 192 & (AR=1)  
& 0.597 & 0.551 & 0.410 & 0.434 & 0.619 & 0.566 & 0.658 & 0.597 \\
& 384 & (AR=2)  
& 0.665 & 0.574 & 0.497 & 0.480 & 0.708 & 0.609 & 0.693 & 0.613 \\
& 768 & (AR=4)  
& 0.802 & 0.651 & 0.604 & 0.532 & 0.799 & 0.651 & 0.736 & 0.633 \\ \midrule

\multirow{3}{*}{Weather\_336} & 336 & Baseline 
& 0.689 & 0.594 & 0.578 & 0.523 & 0.309 & 0.329 & 0.698 & 0.600 \\
& 336 & (AR=1)  
& 0.663 & 0.554 & 0.567 & 0.514 & 0.712 & 0.604 & 0.617 & 0.564 \\
& 672 & (AR=2)  
& 0.837 & 0.675 & 0.633 & 0.577 & 0.854 & 0.669 & 0.633 & 0.577 \\ \midrule

\multirow{2}{*}{Weather\_720} & 720 & Baseline 
& 0.926 & 0.710 & 1.059 & 0.741 & 0.404 & 0.385 & 0.930 & 0.711 \\
& 720 & (AR=1)  
& 0.908 & 0.682 & 0.776 & 0.646 & 0.702 & 0.619 & 0.776 & 0.646 \\

\bottomrule
\end{tabular}
\end{table*}

We now consider the efficiency of AROpt for other models. Before proceeding, we first reintroduce the data structure used by these models. As shown in Fig.~\ref{fig_pipeline}a, only the iTransformer variants have non-overlapping input-output windows and require only the historical input values and their corresponding timestamps during inference. In contrast, most Transformer-based time-series forecasting models require both encoder and decoder inputs during inference. As illustrated in the following data structure, the encoder receives the historical input values together with their corresponding timestamps. The decoder input consists of two parts: the known historical values and zero padding of prediction length \(T\) \citep{liu2023itransformer}. Similarly, the decoder timestamps contain both the timestamps corresponding to the known historical decoder inputs and the timestamps of the future prediction horizon.

\begin{figure*}[ht]
  \begin{center}
    \centering
    \includegraphics[width=\textwidth]{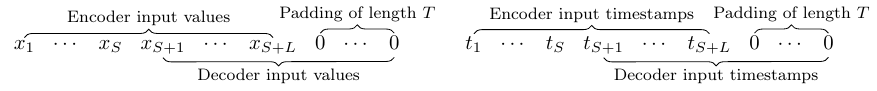}
  \end{center}
\end{figure*}

Table~\ref{tab:sr} shows the robustness of AROpt on the non-iTransformer family, consistent with the main results in Table~\ref{tab:ar}. AROpt enhances small models to achieve advanced performance in most short- and long-term forecasting, but does not outperform the baseline on short-term Weather forecasting with Flowformer and Flashformer at short prediction horizons. 

However, there is a significant difference between the two sets of experiments. Since the overlapping window (see Section~\ref{sec:algorithm}) in the non-iTransformer family has a LARGE non-zero length (48), which introduces partial ground-truth information into the MSE evaluation and affects its trend.

To further evaluate the robustness of AROpt, Table~\ref{tab:rb} presents results on additional datasets using the iTransformer, following the convention of \citep{liu2023itransformer}. The table demonstrates the robustness of our method and its performance gains across diverse domains. For fairness, we do not conduct experiments on other models for these datasets because the reference papers do not provide the corresponding training recipes.

Remarkably, we do not modify any model hyperparameters or retrain the baselines; all baseline results are directly taken from the published papers. For example, setting the projector size of the baseline models to 768 would allow a more convenient comparison with our AROpt-trained models. However, we do \textbf{NOT} do this, because model performance depends heavily on the training recipes. If the reference paper does not provide such a recipe, we cannot design one ourselves and then claim that “this is the baseline, and our method is better,” as this would constitute an unfair comparison. Instead, we compare our results with a sequence length of 768 to baseline results with a length of 720, even though this setting is less favorable to our method.

\begin{table}[h]
\caption{Forecasting performance of iTransformer on additional datasets. ``AR=$k$" indicates $k$-step AR rollout from a short-horizon forecasting model.
}
\label{tab:rb}
\centering
\resizebox{\textwidth}{!}{
\begin{tabular}{l|cc|cc|cc|cc|cc|cc}
\toprule
& \multicolumn{2}{c|}{PEMS04} 
& \multicolumn{2}{c|}{PEMS07} 
& \multicolumn{2}{c|}{Solar} 
& \multicolumn{2}{c|}{Exchange} 
& \multicolumn{2}{c|}{ETTh1} 
& \multicolumn{2}{c}{ETTh2} \\
\midrule
Pred. Len. 
& MSE & MAE & MSE & MAE & MSE & MAE 
& MSE & MAE & MSE & MAE & MSE & MAE \\
\midrule

Baseline 
& 0.150 & 0.262 
& 0.139 & 0.245 
& 0.203 & 0.239 
& {\color{blue}0.086} & 0.206 
& 0.386 & 0.405 
& 0.297 & 0.349 \\

96 (AR=1) 
& {\color{blue}0.136} & {\color{red}0.242} 
& {\color{blue}0.125} & {\color{red}0.224} 
& {\color{blue}0.174} & {\color{red}0.218} 
& 0.099 & {\color{red}0.186} 
& {\color{blue}0.382} & {\color{red}0.398} 
& {\color{blue}0.255} & {\color{red}0.322} \\

192 (AR=2) 
& 0.172 & 0.274 
& 0.162 & 0.255 
& 0.226 & 0.256 
& 0.274 & 0.306 
& 0.412 & 0.418 
& 0.315 & 0.360 \\

288 (AR=3) 
& 0.181 & 0.282 
& 0.173 & 0.261 
& 0.248 & 0.273 
& 0.403 & 0.398 
& 0.431 & 0.433 
& 0.352 & 0.388 \\

384 (AR=4) 
& 0.196 & 0.293 
& 0.186 & 0.270 
& 0.267 & 0.287 
& 0.478 & 0.441 
& 0.436 & 0.442 
& 0.370 & 0.404 \\

\bottomrule
\end{tabular}
}
\end{table}

\begin{figure}[h]
  \begin{center}
    \centering
    \includegraphics[width=\textwidth]{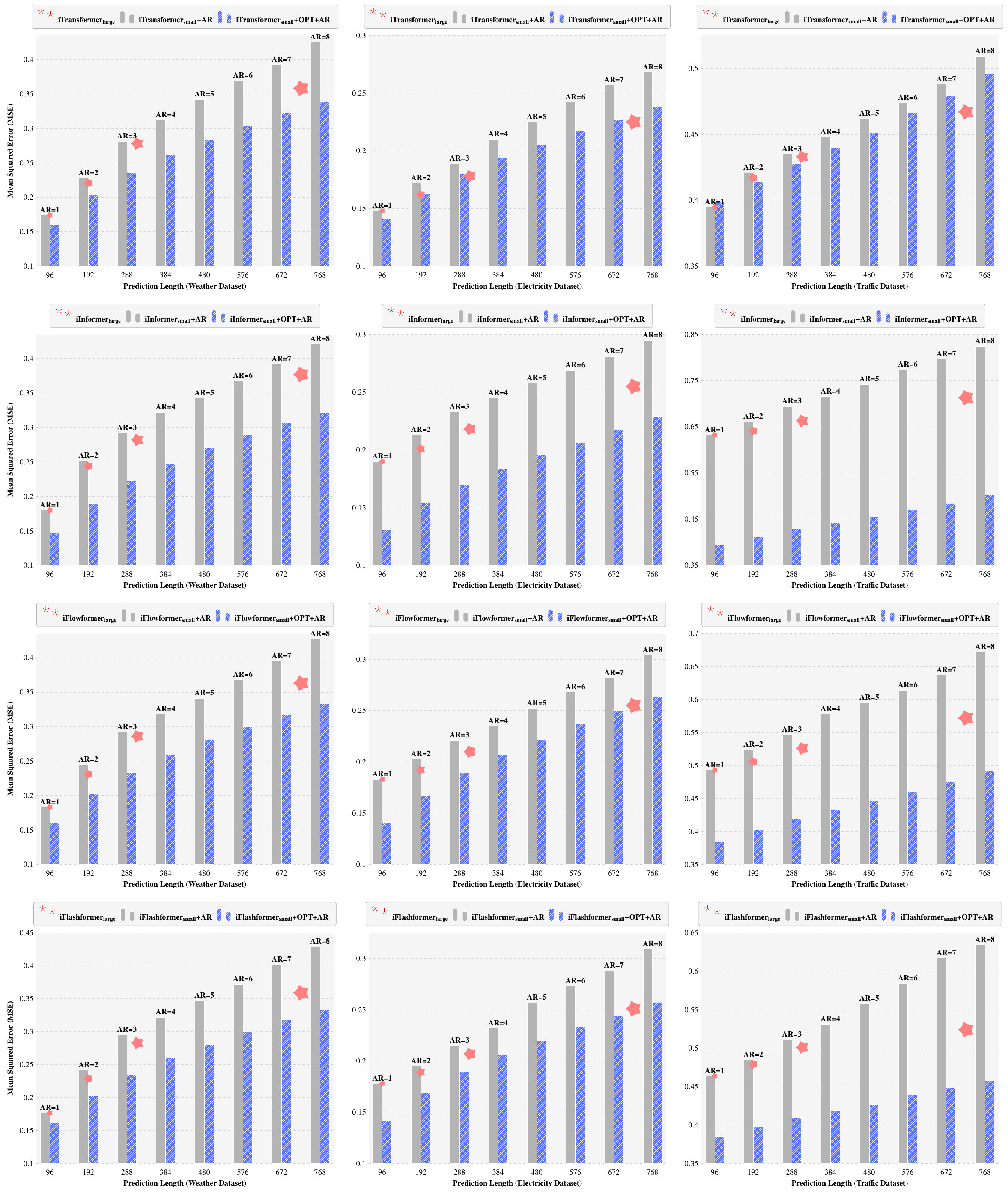}
    \caption{
    Forecasting performance on the Electricity, Traffic, and Weather datasets with lookback length $S = 96$. Red stars denote scaled large models, while bars represent the same fixed-size small model, with blue indicating training using our proposed optimization method.
    }
    \label{fig_ablation_study_full}
  \end{center}
\end{figure}

\section{Ablation Study}
\label{appendix:Ablation Study}

Fig. \ref{fig_ablation_study_full} presents the ablation studies across multiple models (iTransformer, iInformer, iFlowerformer, and iFlashformer) and datasets (Weather, Electricity, and Traffic). Red stars, with increasing marker size, indicate a sequence of scaled-up models with projector sizes of 96, 192, 336, and 720, and the MSE values are taken from the iTransformer paper. Gray bars represent a fixed-size model with projector dimension 96, where long-horizon predictions are obtained by concatenating short-term predictions (i.e., 96 × AR steps). Blue bars denote the same small model trained with our proposed optimization method under AR rollout.

These results further reinforce our conclusion in Section \ref{sec:Results}: while scaling up the model improves performance over applying AR rollout to a short-horizon forecasting model, it remains inferior to applying AR rollout to a short-horizon forecasting model trained with our proposed optimization method.

\section{Sensitivity Analysis}
\label{appendix:Sensitivity Analysis}

\begin{figure}[ht]
  \vskip 0.2in
  \begin{center}
    \centerline{
        \begin{tikzpicture}
        \node[inner sep=0pt, outer sep=0pt] (img) {%
          \includegraphics[
            width=\columnwidth,
            trim=1cm 46.25cm 1cm 1.5cm,
            clip
          ]{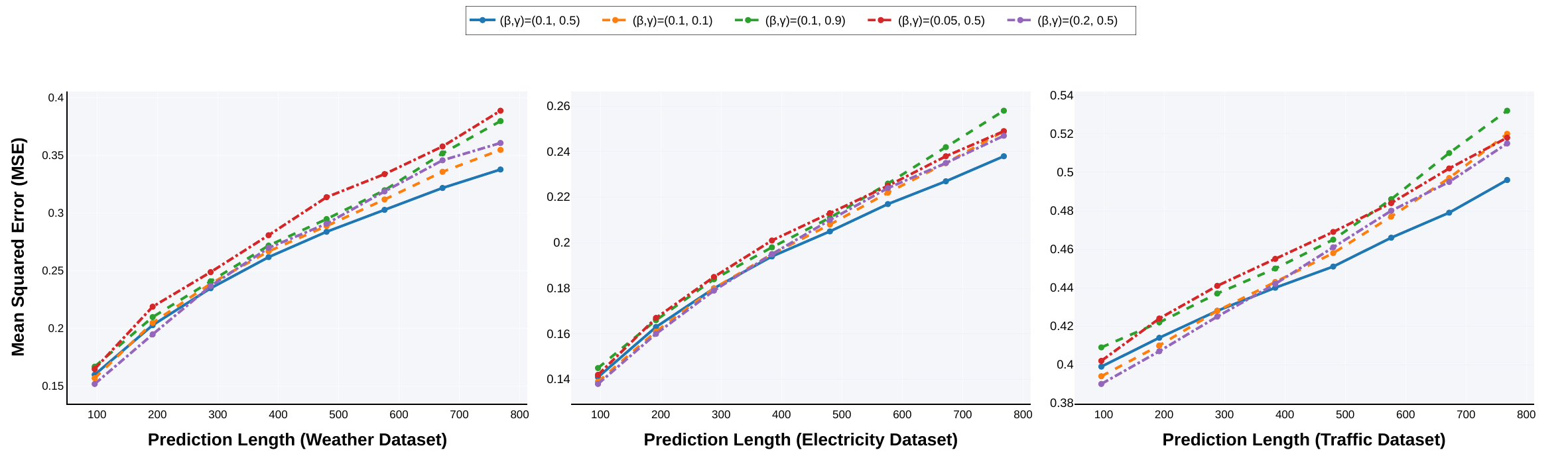}%
        };
        \end{tikzpicture}
    }
    \caption{Hyperparameter sensitivity with respect to $\beta$ and $\gamma$ on the Weather, Electricity, and Traffic datasets using iTransformer, where the input and output lengths are both 96. Each subplot shows eight AR results \((\text{AR} = 1, \dots, 8)\), corresponding to prediction lengths of $96, \dots, 96 \times 8$, under five $(\beta, \gamma)$ configurations.}
    \label{fig_sensitivity_analysis}
  \end{center}
\end{figure}
Based on our theoretical analysis in Section \ref{sec:algorithm}, the hyperparameter $\gamma$ should be set to 0.5 so that the gradient norm of our loss is aligned in scale with that of the traditional MSE loss (Eq. \ref{eq:estimation}). From Definition (\ref{eqn:penalty}), if $\beta = 0$, we have
\[
\nabla_{\theta} r_k = -\nabla_{\theta} e_k,\quad \nabla_{\theta} \ell = \sum_{k=0}^{n-1} \gamma^k \nabla_{\theta} e_k.
\]
This implies that the gradient of our loss $\ell$ reduces to a weighted sum of the gradients of the standard MSE terms. If $\beta = 0.5$, we have
\[
\nabla_{\theta} r_k = 0, \quad \text{if } e_k < e_{k-1},
\]
which means that our loss $\ell$ depends only on those patches where rollout inconsistencies do not occur. Therefore, we limit $\beta \in (0, 0.5)$.

From an experimental perspective, Fig.~\ref{fig_sensitivity_analysis} illustrates the sensitivity of our optimization method with respect to the hyperparameters $\beta$ and $\gamma$. We find that the default setting $(\beta, \gamma) = (0.1, 0.5)$ is empirically chosen, as it achieves relatively low MSE in short-horizon forecasting and the lowest MSE in long-horizon forecasting.

\section{Analysis of Early-Horizon Error Prioritization in the One-Step AR Rollout}
\label{app:Analysis of Early-Horizon Error Prioritization in the One-Step AR}
For a one AR rollout step loss,
\[
\ell=-r_0-\gamma r_1,
\qquad 0<\gamma<1,
\]
we have
\[
\nabla_\theta \ell
=
\nabla_\theta e_0
+
\gamma c\nabla_\theta e_1, \qquad
c\in\{1,1-\beta,1-2\beta\}.
\]
Consider the Stochastic Gradient Descent (SGD) update with a small learning rate \(\eta\):
\[
\theta^+=\theta-\eta\nabla_\theta \ell(\theta).
\]
Using a first-order Taylor expansion,
\begin{equation}
e_i(\theta)-e_i(\theta^+)
=
\eta
\left\langle
\nabla_\theta e_i,
\nabla_\theta \ell
\right\rangle
+O(\eta^2),
\qquad i\in\{0,1\}.
\label{eq:Taylor expansion}
\end{equation}

Assume
\begin{equation}
e_0>e_1
\quad\Longrightarrow\quad
\|\nabla_\theta e_0\|\ge\|\nabla_\theta e_1\|.
\label{eq:Assume}
\end{equation}
This assumption is natural, since the loss component associated with the larger error is expected to generate at least as strong a local optimization signal. Using the Cauchy--Schwarz inequality together with Eq.~(\ref{eq:Taylor expansion}) and inequality~(\ref{eq:Assume}),
\[
\begin{aligned}
&
[e_0(\theta)-e_0(\theta^+)] - [e_1(\theta)-e_1(\theta^+)] \\
& = \eta (\langle \nabla_\theta e_0,\nabla_\theta\ell\rangle - \langle \nabla_\theta e_1,\nabla_\theta\ell\rangle) + O(\eta^2) \\
& \approx \eta (\langle \nabla_\theta e_0,\nabla_\theta\ell\rangle - \langle \nabla_\theta e_1,\nabla_\theta\ell\rangle) \\
&=
\eta \|\nabla_\theta e_0\|^2 - \eta \gamma c\|\nabla_\theta e_1\|^2 - \eta (1-\gamma c)\langle \nabla_\theta e_0,\nabla_\theta e_1\rangle\\
&\ge \eta (\|\nabla_\theta e_0\|-\|\nabla_\theta e_1\|) (\|\nabla_\theta e_0\|+\gamma c\|\nabla_\theta e_1\|)\\
&\ge 0.
\end{aligned}
\]
Therefore, SGD reduces the earlier and larger error at least as strongly as the discounted later error. 

\section{Asymptotic and Non-asymptotic Analysis}
\label{appendix:Asymptotic and Non-asymptotic Analysis}
Based on Eqs. (\ref{eqn:penalty}) and (\ref{eqn:loss}), the gradient of our loss $\ell$ is:

\[
\nabla \ell = \nabla \ell_1 + \nabla \ell_2 + \nabla \ell_3,
\] 

where 

\[
\begin{cases}
\nabla \ell_1 &= \nabla \frac{1}{T} \sum\limits_{t=S}^{S+T-1} |x_t - \hat{x}_t|^2 \\
&= \nabla \operatorname{MSE}(X_{S:S+T}; \hat{X}_{S:S+T}), \\
\nabla \ell_2 &= \sum\limits_{k=1}^{n-1} (1-\beta) \gamma^k \nabla \frac{1}{T} \sum\limits_{t=S+kT}^{S+(k+1)T-1} |x_t - \hat{x}_t|^2 \\
&= \sum\limits_{k=1}^{n-1} (1-\beta) \gamma^k \nabla \operatorname{MSE}(X_{S:S+kT}; \hat{X}_{S:S+(k+1)T}), \\
\nabla \ell_3 &= \sum\limits_{k=1}^{n-1} \operatorname{sgn}(e_k-e_{k-1}) \beta \gamma^k \nabla \frac{1}{T} \sum\limits_{t=S+kT}^{S+(k+1)T-1} |x_t - \hat{x}_t|^2 \\
&= \sum\limits_{k=1}^{n-1} \operatorname{sgn}(e_k-e_{k-1}) \beta \gamma^k \nabla \operatorname{MSE}(X_{S:S+kT}; \hat{X}_{S:S+(k+1)T}). 
\end{cases}
\]

This demonstrates that our $\nabla \ell$ is a linear combination of standard MSE gradients.  

\begin{itemize}
    \item If $e_k > e_{k-1}$ holds consistently, then $\operatorname{sgn}(e_k-e_{k-1})=1$. In this case, the gradient 
    \[
    \nabla \ell = \sum_{k=0}^{n-1}  \gamma^k \nabla \operatorname{MSE}(X_{S+kT:S+(k+1)T}; \hat{X}_{S+kT:S+(k+1)T})
    \] 
    is a Exponential Moving Average(EMA) of the standard $\nabla \text{MSE}$ with a coefficient $\gamma=0.5$. 
    \item If the model predicts a correct output despite receiving an incorrect input (i.e., a random guess) at the $k$-patch, then $e_k <e_{k-1}$. In the case, $\operatorname{sgn}(e_k - e_{k-1}) = -1$. This leads to a lower weight ($1-2\beta$) to $\nabla \operatorname{MSE}(X_{S+kT:S+(k+1)T}; \hat{X}_{S+kT:S+(k+1)T})$ for the $k$-th patch predictions (since $\beta \in (0, 0.5)$, the weight remains greater than 0). 
\end{itemize}

In conclusion,  our $\nabla \ell$ is a linear combination of the standard MSE gradients. If the MSE function of a time-series forecasting model converges during training, our method is theoretically expected to converge as well.

The optimizer employed in our experiments is identical to the Adam optimizer \cite{kingma2014adam} used in the iTransformer GitHub repository. Now, we can make a overall conclusion.

\textbf{Asymptotic Analysis}
\begin{itemize}
    \item In convex MSE problems (e.g., linear regression), Adam can converge to the global minimum, but only under certain conditions, for example, the single linear layer network.
    \item In non-convex settings (deep neural networks), Adam converges to a stationary point, not necessarily the best minimum. [\cite{zhang2022adam}]
    \item Its adaptive updates may lead to different solutions compared to SGD.
\end{itemize}

\textbf{Non-Asymptotic Analysis}
\begin{itemize}
    \item Adam achieves a convergence rate of roughly $\mathcal{O}(1/\sqrt{T})$ for MSE \cite{chen2019convergence}.
    \item But in most cases, it often shows (1) faster initial error reduction; (2) better stability with noisy gradients; (3) less sensitivity to scaling \cite{kingma2014adam}.
\end{itemize}

\section{Comparison with Time-Series Foundation Models}
\label{app:Comparison with Time-Series Foundation Models}

Recent time-series foundation models (TSFMs), including Chronos \cite{ansari2024chronos},
Moirai \cite{woo2024unified}, TimesFM \cite{das2023decoder}, and Timer \cite{liu2024timer},
represent an important direction for time-series forecasting.
However, these methods are developed under a substantially different training
paradigm from AROpt. As summarized in Table~\ref{tab:tsfm}, TSFMs are pretrained
on large-scale corpora and are primarily evaluated in zero-shot, few-shot, or
fine-tuning settings, whereas AROpt is a supervised optimization strategy for
improving autoregressive forecasting models under the standard long-horizon
forecasting protocol.

\begin{table}[h]
\centering
\small
\caption{Comparison between recent time-series foundation models and AROpt.}
\label{tab:tsfm}
\begin{tabular}{p{1.8cm}p{2cm}p{3cm}p{5cm}}
\toprule
Method & TParadigm & Primary evaluation & Remark \\
\midrule
Chronos~\cite{ansari2024chronos}
& Large-scale pretrained TSFM
& Zero-shot / Fine-tuning
& Mainly evaluated on short-horizon forecasting (prediction length 4--56). \\
\addlinespace

Moirai~\cite{woo2024unified}
& Universal pretrained TSFM
& Zero-shot / Few-shot / Fine-tuning
& Reports average MSE (Weather 0.238–0.250, Electricity 0.188–0.233, Table 6) rather than MSE across prediction lengths; "Moirai + full fine-tuning" remains below "iInformer + AROpt" on benchmarks. \\
\addlinespace

TimesFM~\cite{das2023decoder}
& Decoder-only TSFM
& Zero-shot / Fine-tuning
& ETTh1 0.398–0.445, ETTh2 0.356–0.457 (Table 2); "TimesFM + fine-tuning" remains below "iTransformer + AROpt" on benchmarks. \\
\addlinespace

Timer~\cite{liu2024timer}
& Generative pretrained TSFM
& Zero-shot / Fine-tuning
& Uses longer 672-step input while mainly evaluating 96-step prediction (Electricity 0.132, Traffic 0.351, Weather 0.154, Table 10); "Timer + full fine-tuning" remains below "iInformer + AROpt" on benchmarks. \\
\addlinespace

\textbf{AROpt (ours)}
& Supervised optimization method
& Fully supervised training
& Standard long-horizon forecasting (input length $96$, prediction length $96$--$720$). \\
\bottomrule
\end{tabular}
\end{table}

Although these foundation models can also be fully fine-tuned on downstream datasets, their training paradigms and evaluation protocols differ substantially from the fully supervised long-horizon forecasting setting considered in this work. In particular, TSFMs benefit from large-scale pretraining on substantially more time-series data before downstream fine-tuning, whereas AROpt is trained solely on the target benchmark datasets without any pretraining. Consequently, a direct numerical comparison is not entirely fair and, if anything, is more favorable to the pretrained foundation models. Nevertheless, the reported fully fine-tuned results of recent TSFMs on common benchmark datasets (e.g., Weather, Electricity, ETTh1, and ETTh2) still do not consistently surpass the performance achieved by AROpt under the standard evaluation protocol (input length =96, prediction lengths =96--720). Therefore, we compare AROpt with state-of-the-art supervised long-horizon forecasting methods that follow the same experimental protocol, which we believe provides the fairest assessment of the proposed optimization strategy.

\section{Runtime Analysis}
\label{appendix:runtime}

In this appendix, we analyze the computational overhead of AROpt during both training and inference. Since all experiments employ early stopping, the total training time varies across datasets and models. Therefore, we report the average training runtime per epoch, average inference runtime per batch, and peak GPU memory usage instead. All measurements are obtained on the Weather dataset using an NVIDIA RTX A6000 GPU with a batch size of 512.

\begin{table}[h]
\centering
\small
\caption{Training and inference overhead under different AR rollout depths.}
\label{tab:runtime}
\begin{tabular}{cp{2cm}p{2cm}p{2cm}p{2cm}}
\toprule
AR Steps & Train Runtime / Epoch (s) & Train Peak Memory (MB) & Inference Runtime / Batch (s) & Inference Peak Memory (MB) \\
\midrule
1 & 3.57  & 1332.57 & 0.0024 & 67.85 \\
2 & 6.79  & 2337.41 & 0.0034 & 70.99 \\
4 & 13.17 & 4347.58 & 0.0051 & 70.02 \\
8 & 25.70 & 8365.02 & 0.0089 & 71.71 \\
\bottomrule
\end{tabular}
\end{table}

As expected, both the training runtime and GPU memory usage increase approximately linearly with the AR rollout depth. However, this computational growth is an inherent property of autoregressive forecasting rather than a limitation introduced by AROpt. The proposed method neither modifies the model architecture nor increases the computational complexity of each forward or backward pass beyond the standard AR rollout procedure.

\begin{table*}[!b]
\centering
\small
\caption{Multi-seed evaluation on the Weather dataset. Results are reported as mean $\pm$ standard deviation over five independent runs.}
\label{tab:multiseed}
\begin{tabular}{cc|cccc}
\toprule
\multicolumn{6}{c}{\textbf{MSE}}\\
\midrule
Pred.\ Len. & AR
& iTransformer
& iInformer
& iFlowformer
& iFlashformer\\
\midrule
96  & AR=1 & $0.163_{\pm0.004}$ & $0.148_{\pm0.002}$ & $0.162_{\pm0.001}$ & $0.160_{\pm0.001}$\\
192 & AR=2 & $0.205_{\pm0.003}$ & $0.191_{\pm0.002}$ & $0.203_{\pm0.000}$ & $0.202_{\pm0.001}$\\
384 & AR=4 & $0.263_{\pm0.001}$ & $0.249_{\pm0.002}$ & $0.259_{\pm0.001}$ & $0.261_{\pm0.001}$\\
768 & AR=8 & $0.338_{\pm0.001}$ & $0.325_{\pm0.005}$ & $0.332_{\pm0.002}$ & $0.336_{\pm0.002}$\\
\midrule
192 & AR=1 & $0.206_{\pm0.003}$ & $0.195_{\pm0.001}$ & $0.204_{\pm0.000}$ & $0.207_{\pm0.003}$\\
384 & AR=2 & $0.262_{\pm0.002}$ & $0.252_{\pm0.002}$ & $0.260_{\pm0.001}$ & $0.263_{\pm0.002}$\\
768 & AR=4 & $0.335_{\pm0.001}$ & $0.327_{\pm0.001}$ & $0.332_{\pm0.001}$ & $0.337_{\pm0.002}$\\
\midrule
336 & AR=1 & $0.252_{\pm0.001}$ & $0.245_{\pm0.001}$ & $0.251_{\pm0.003}$ & $0.252_{\pm0.002}$\\
672 & AR=2 & $0.320_{\pm0.000}$ & $0.317_{\pm0.002}$ & $0.319_{\pm0.003}$ & $0.321_{\pm0.002}$\\
\midrule
720 & AR=1 & $0.330_{\pm0.001}$ & $0.324_{\pm0.002}$ & $0.327_{\pm0.002}$ & $0.328_{\pm0.001}$\\

\midrule
\multicolumn{6}{c}{\textbf{MAE}}\\
\midrule
Pred.\ Len. & AR
& iTransformer
& iInformer
& iFlowformer
& iFlashformer\\
\midrule
96  & AR=1 & $0.212_{\pm0.004}$ & $0.202_{\pm0.001}$ & $0.211_{\pm0.000}$ & $0.209_{\pm0.002}$\\
192 & AR=2 & $0.244_{\pm0.013}$ & $0.243_{\pm0.001}$ & $0.249_{\pm0.001}$ & $0.249_{\pm0.001}$\\
384 & AR=4 & $0.294_{\pm0.001}$ & $0.287_{\pm0.002}$ & $0.291_{\pm0.001}$ & $0.293_{\pm0.001}$\\
768 & AR=8 & $0.342_{\pm0.000}$ & $0.336_{\pm0.004}$ & $0.337_{\pm0.002}$ & $0.336_{\pm0.013}$\\
\midrule
192 & AR=1 & $0.252_{\pm0.001}$ & $0.246_{\pm0.000}$ & $0.251_{\pm0.000}$ & $0.252_{\pm0.002}$\\
384 & AR=2 & $0.295_{\pm0.000}$ & $0.290_{\pm0.000}$ & $0.294_{\pm0.001}$ & $0.295_{\pm0.002}$\\
768 & AR=4 & $0.342_{\pm0.001}$ & $0.337_{\pm0.000}$ & $0.340_{\pm0.000}$ & $0.343_{\pm0.002}$\\
\midrule
336 & AR=1 & $0.287_{\pm0.002}$ & $0.285_{\pm0.001}$ & $0.287_{\pm0.002}$ & $0.288_{\pm0.001}$\\
672 & AR=2 & $0.332_{\pm0.001}$ & $0.332_{\pm0.000}$ & $0.331_{\pm0.002}$ & $0.333_{\pm0.001}$\\
\midrule
720 & AR=1 & $0.338_{\pm0.001}$ & $0.337_{\pm0.002}$ & $0.337_{\pm0.003}$ & $0.336_{\pm0.003}$\\
\bottomrule
\end{tabular}
\end{table*}

Moreover, the practical overhead remains small. Even with eight rollout steps, the average training time is less than 30 seconds per epoch, while the inference latency is only $0.0024/512$--$0.0089/512$ seconds per sample. These results indicate that AROpt remains computationally efficient for long-horizon forecasting. Therefore, we attribute the observed linear computational growth to the autoregressive forecasting paradigm itself, rather than to the proposed optimization method.

\section{Multi-Seed Evaluation}
\label{appendix:multiseed}

In this appendix, we provide additional multi-seed experiments to evaluate the statistical stability of AROpt. While the main paper follows the original implementation and adopts a fixed random seed ($2023$) for fair paired comparison with previous work, we additionally repeat the complete training pipeline using five different random seeds and report the results as $\mathrm{mean}\pm\mathrm{std}$.

To reproduce these experiments, the fixed random seed in \texttt{run.py} should be disabled by commenting out

\begin{verbatim}
# fix_seed = 2023
# random.seed(fix_seed)
# torch.manual_seed(fix_seed)
# np.random.seed(fix_seed)
\end{verbatim}

The script \texttt{weather.sh} is then executed five independent times. Overall, this evaluation corresponds to training $16\times5$ models and performing $80\times5$ evaluations on the Weather benchmark. The resulting performance is summarized in Table~\ref{tab:multiseed}.

As shown in Table~\ref{tab:multiseed}, the standard deviations remain consistently small across different prediction horizons, autoregressive rollout settings, and backbone architectures. For most configurations, the standard deviation is below $0.005$ for both MSE and MAE, indicating that AROpt exhibits stable optimization behavior under different random initializations. These results further support that the performance improvements reported in the main paper are robust and are unlikely to be explained by a favorable random seed alone.

For complete transparency and reproducibility, we will release the complete training logs and evaluation outputs through the anonymous supplementary repository in accordance with the conference artifact policy.



\end{document}